\documentclass[letterpaper]{article} 
\usepackage{aaai23}  
\usepackage{times}  
\usepackage{helvet}  
\usepackage{courier}  
\usepackage[hyphens]{url}  
\usepackage{graphicx} 
\usepackage{natbib}  
\setcitestyle{square}
\usepackage{caption} 
\usepackage{algorithm}
\usepackage{algorithmic}

\usepackage{newfloat}
\usepackage{listings}
\DeclareCaptionStyle{ruled}{labelfont=normalfont,labelsep=colon,strut=off} 
\floatstyle{ruled}
\newfloat{listing}{tb}{lst}{}
\floatname{listing}{Listing}

\usepackage{graphicx}
\usepackage{amsfonts}
\usepackage{amsmath}
\usepackage{amssymb}
\usepackage{amsthm}
\usepackage{multirow}
\newcommand\scalemath[2]{\scalebox{#1}{\mbox{\ensuremath{\displaystyle #2}}}}
\usepackage{hyperref}
\usepackage{booktabs}
\usepackage{balance}

\title{RePreM: Representation Pre-training with Masked Model for \\Reinforcement Learning}
\author{
    Yuanying Cai\textsuperscript{\rm 1},
    Chuheng Zhang\textsuperscript{\rm 2},
    Wei Shen\textsuperscript{\rm 3},
    Xuyun Zhang\textsuperscript{\rm 4},
    Wenjie Ruan\textsuperscript{\rm 4,5},
    Longbo Huang\textsuperscript{\rm 1}\thanks{Corresponding Author}
}
\affiliations{
\textsuperscript{\rm 1} IIIS, Tsinghua University, Beijing, China \qquad
\textsuperscript{\rm 2} Microsoft Research Asia, Beijing, China\\
\textsuperscript{\rm 3} Hulu, Beijing, China \qquad
\textsuperscript{\rm 4} Macquarie University, Sydney, Australia   \qquad
\textsuperscript{\rm 5} University of Exeter, Exeter, UK\\
cai-yy16@mails.tsinghua.edu.cn,
chuhengzhang@microsoft.com,
wei.shen@disneystreaming.com, \\
xuyun.zhang@mq.edu.au, 
wruan@mq.edu.au,
longbohuang@tsinghua.edu.cn
}

\usepackage{bibentry}

\begin{document}

\maketitle

\begin{abstract}
    
Inspired by the recent success of
sequence modeling in RL 
and the use of masked language model for pre-training, 
we propose a masked model for pre-training in RL,
RePreM (\textbf{Re}presentation \textbf{Pre}-training with \textbf{M}asked Model), which trains the encoder combined with transformer blocks to predict the masked states or actions in a trajectory.
RePreM is simple but effective compared to existing representation pre-training methods in RL.
It avoids algorithmic sophistication (such as data augmentation or estimating multiple models) with sequence modeling 
and generates a representation that captures long-term dynamics well.
Empirically, we demonstrate the effectiveness of RePreM in various tasks, including dynamic prediction, transfer learning, and sample-efficient RL with both value-based and actor-critic methods.
Moreover, we show that RePreM scales well with dataset size,  dataset quality, and the scale of the encoder, 
which indicates its potential towards big RL models.

\end{abstract}

\vspace{-10pt}

\section{Introduction}
\label{sec:introduction}

Reinforcement learning (RL) has achieved great success in solving various decision-making tasks \citep{mnih2015human,silver2016mastering,arulkumaran2019alphastar,jumper2021highly}. Yet, the sample inefficiency issue \citep{tsividis2017human} and the overfitting problem \citep{zhang2018study} impede its further application to real-world scenarios.
In contrast, humans can typically learn more efficiently and robustly than current RL algorithms.
The gap is resulted from the fact that existing RL algorithms usually start learning \emph{tabula rasa} without any prior knowledge \citep{schwarzer2021pretraining, hutsebaut2020pre}. 
As a result, it requires a large number of samples to learn meaningful representations \citep{dubey2018investigating,lake2017building}.

In this paper, we consider learning state representations that incorporate rich knowledge by unsupervised pre-training from offline datasets.
While unsupervised pre-training methods have achieved great success in natural language processing~(NLP) \citep{devlin2018bert,brown2020language} and  computer vision~(CV) \citep{henaff2020data,he2020momentum,grill2020bootstrap,chen2020simple}, they are relatively underexplored in RL.
The main benefit of pre-trained representations is that they can be used for sample-efficient learning for various downstream tasks such as dynamics prediction, transfer learning, and reinforcement learning with different algorithms and different reward functions.

However, existing methods that generate pre-trained representations suffer from the following limitations:
1) They are algorithmically complex.  
Some of these methods, especially those based on self-supervised learning, rely on hand-crafted data augmentation that needs specific domain knowledge \citep[see e.g.,][]{stooke2021decoupling,schwarzer2021pretraining}, while others are based on the world model and require additional networks to model latent transition or reward functions \citep[see e.g.,][]{seo2022reinforcement}.
2) We find that, although several techniques, such as goal-conditioned RL \citep{schwarzer2021pretraining} and RSSM \citep{hafner2019learning}, are designed to encourage representations to incorporate the information over the multi-step dynamics in the future, previous methods cannot predict long-term dynamics accurately, which limits their performance (see our experiments in Section \ref{sec:dynamic_prediction}).

Recently, sequence modeling (mostly with Transformer) has achieved great success in RL \citep{chen2021decision,janner2021offline}. 
This approach can avoid algorithmic sophistication by following the RvS (RL via supervised learning) paradigm \citep{emmons2021rvs} and extract long-term information more effectively by learning the whole sequence.
However, most previous work on sequence modeling for RL focuses on the offline RL setting \citep{chen2021decision,janner2021offline,reed2022generalist,furuta2021generalized,akhmetov2022bootstrapping,xu2022prompting}, where the agent's performance is limited by the quality of the offline datasets.
There are two exceptions \citep{zheng2022online, lee2022multi} that 
fine-tune the models pre-trained from offline datasets.
However, these two methods use the full pre-trained model instead of the representation, which limits the variety of applicable downstream tasks.

To effectively learn presentations that offer flexibility to downstreaming RL tasks, 
we propose RePreM (\textbf{Re}presentation \textbf{Pre}-training with \textbf{M}asked Model) that pre-trains the encoder to predict the masked states or actions in the offline trajectories with the help of transformer blocks \citep{vaswani2017attention}. 
Specifically, we treat the trajectory $(s_1, a_1, s_2, a_2, \cdots)$ as a sequence and randomly mask a proportion of states or actions in the sequence.
Then, we generate the representations of the states/actions in the sequence with the corresponding encoder and feed these representations to the transformer blocks.
We train the encoders and the transformer blocks to predict the embedding of the masked states/actions with a contrastive loss.
Predicting the masked states and actions with a uniform encoder-transformer architecture nicely corresponds to the role of the forward and inverse transition models designed in previous pretraining tasks (see, e.g., \citep{yang2021representation,schwarzer2021pretraining}).
When a specific downstream task is given, we build the corresponding networks (e.g., the policy or the value network) simply by appending several MLP layers to the fixed state encoder.
In this way, we can achieve superior performance on various of tasks sample-efficiently.

RePreM is motivated by the repeated success achieved by similar architectures (i.e., transformer blocks combined with the masked model) in NLP \citep{devlin2018bert,song2019mass} and CV \citep{carion2020end,dosovitskiy2020image} for sequential information processing where long-range data dependency matters, a successful example of which is BERT \citep{devlin2018bert}.
Compared to previous methods that require data augmentation or modeling the transition/reward process explicitly, RePreM is simple since it learns the state encoder with only
transformer blocks and a self-supervision loss that predicts the masked elements.
Furthermore, with the help of 
sequence modeling,
the pre-trained encoder by RePreM can capture the long-term dynamics effectively.

We conduct extensive experiments on Atari games \citep{bellemare2013arcade} and DeepMind Control Suite (DMControl) \citep{tassa2018deepmind}. 
We show that our pre-trained state encoder enables sample-efficient learning on several downstream tasks including dynamic prediction, transfer learning, and sample-efficient RL.
For dynamic prediction, our encoder results in a smaller prediction error than the baselines, especially for long-horizon predictions, which demonstrates that RePreM can capture the long-term dynamics effectively.
For transfer learning, we pre-train the encoder on the data from a set of 24 Atari games and successfully transfer the representation to unseen games.
For sample-efficient RL, we evaluate the pre-trained representation with the 100k benchmark (where the agent is only allowed to interact with the environment for 100k steps) proposed by \citet{kaiser2019model}.
The results show that the representation learned by RePreM boosts Rainbow (a value-based method) \citep{hessel2018rainbow} and SAC (an actor-critic method) \citep{haarnoja2018soft}, and achieves superior performance in Atari and DMControl respectively.
Moreover, we show that the performance of RePreM scales well with the volume of the dataset, the quality of the dataset, and the size of the encoder.
This indicates that this BERT-style pre-training method has great potential in learning representations to extract rich prior knowledge from big data.
Finally, we conduct ablation studies on different designs of RePreM.

The contributions of our paper are summarized as follows:
\begin{itemize}
    \item We propose a simple yet effective pre-training method called RePreM (Representation Pre-training with Masked Model) that incorporates rich information and captures long-term dynamics for the representations in RL. To the best of our knowledge, we are the first to adopt BERT-style representation pre-training in RL.
    \item We conduct extensive experiments and show that RePreM can generate representations that enable sample-efficient learning in various downstream tasks.
    Specifically, we achieve state-of-the-art performance on the Atari-100k benchmark.
    Moreover, we show that its superior performance comes from its ability to capture long-term dynamics and it scales well with the dataset volume/quality and the encoder size.
\end{itemize}

\section{Related Work}
\label{sec:related_work}

\begin{figure*}[tb]
   \centering
   \includegraphics[width=1.7\columnwidth]{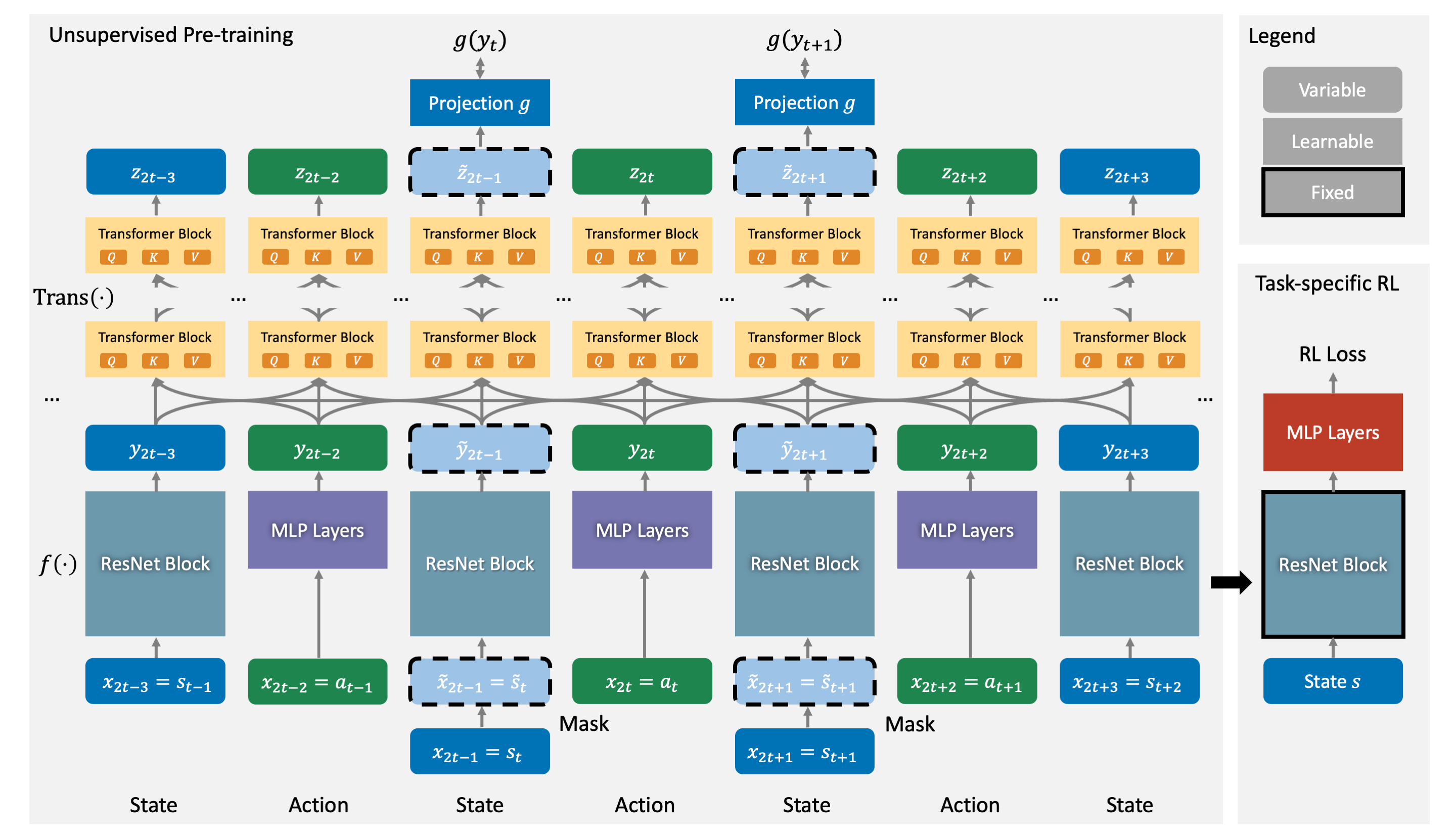}
   \caption{
   Unsupervised pre-training phase and the task-specific reinforcement learning phase of our method.
   }
\label{fig:model}
\end{figure*}

\textbf{Representation learning in RL.}
The studies on representation learning in RL can be broadly divided into two categories: representation learning along with online RL (e.g., \citep{schwarzer2020data}) and pre-training representations based on offline data (e.g., \citep{schwarzer2021pretraining}).
The objective of representation learning in RL is to generate generalizable representations that can aggregate similar states.
However, some of the previous methods ignore the relationship between one state and other states or actions in the trajectory.
Therefore, the resultant representations do not contain information about the environmental dynamics.
Examples of this type include reconstruction-based methods \citep{jaderberg2016reinforcement,yarats2019improving} and data augmentation-based methods \citep{bachman2019learning,srinivas2020curl,stooke2021decoupling}.
Other methods incorporate the dynamics to representations with bisimulation/other abstraction techniques \citep{castro2020scalable,zhang2020learning,liu2020return}, prediction-based approaches \citep{gelada2019deepmdp} or Laplacian representations \citep{mahadevan2007proto,wu2018laplacian}.
However, these methods only consider the one-step transition dynamics and therefore cannot incorporate the long-term dynamics into the representation.
To capture long-term dynamics, recent methods adopt techniques such as predicting multiple steps in the future \citep{schwarzer2020data}, goal-conditioned RL \citep{schwarzer2021pretraining}, recurrent architecture \citep{guo2020bootstrap}, or recurrent state space model (RSSM) \citep{hafner2019dream,hafner2019learning,hafner2020mastering}.
Nevertheless, these methods are complex due to the reliance on multiple models and loss functions.
In contrast, our method is simple (only with an additional transformer structure trained by the masked prediction task) and better captures the long-term dynamics.

\noindent\textbf{Sequence Modeling in RL.}
Recently, motivated by the success of sequence modeling in NLP, researchers have begun to adopt it to RL.
Decision Transformer \citep{chen2021decision} and Trajectory Transformer \citep{janner2021offline} first use the transformer structure \citep{vaswani2017attention} to model the RL agent.
However, most of the work focuses on offline RL \citep{reed2022generalist,furuta2021generalized,akhmetov2022bootstrapping,xu2022prompting} or online RL \citep{parisotto2020stabilizing,banino2021coberl}.
Two exceptions are \citet{zheng2022online} and \citet{lee2022multi} that consider the pre-training setting and fine-tune the pre-trained model as the agent/policy later.
Different from these methods that use the pre-trained model, we aim to learn a pre-trained encoder, and this approach has two benefits:
First, pre-trained representation is more portable for various downstream tasks.
Second, since we only take the encoder from the pre-training phase, we can adopt simpler and more efficient bi-directional models instead of autoregressive ones.

\section{Methods}
\label{sec:methods}


In this section, we will introduce how RePreM (Representation Pre-training with Masked Model) trains the state encoder in the pre-training phase and how we use the pre-trained state encoder in downstream RL tasks.

\subsection{Pre-training in RL}

As in many previous papers \citep[e.g.,][]{eysenbach2018diversity,schwarzer2021pretraining}, we consider pre-training from a reward-free dataset of trajectories that include only states and actions.
Compared with pre-training with rewards \citep[e.g.,][]{gelada2019deepmdp}, this setting has two benefits: 1) reward-free datasets are more feasible in practice; and 2) the knowledge extracted from reward-free datasets can be used for different tasks in similar environments.
The objective of this setting is to learn generalizable representations that are independent of the behavior policy (i.e., the policy used to collect the dataset) and the reward function.
In this way, the representation can be trained with the trajectories collected by arbitrary behavior policies and used in tasks with different reward functions.
We denote the offline dataset with $N$ trajectories as $\mathcal{D}=\{(s_{1}^i, a_1^i, s_2^i, a_2^i, \cdots, s_T^i)\}_{i=1}^N$, where $T$ is the length of the trajectory.

Previously, various self-supervised learning tasks have been proposed to extract information from provided agent experiences.
There are two categories of frequently used tasks:
the forward prediction task in which the agent is trained to predict the immediate states \citep{guo2018neural,gelada2019deepmdp,stooke2021decoupling,schwarzer2020data}, and the backward prediction task where the agent is trained to predict the actions that can generate the specified transitions \citep{pathak2017curiosity,shelhamer2016loss,du2019provably}.
Recent success in representation pre-training relies on the combination of different self-supervision tasks \citep{schwarzer2021pretraining}.
Based on the previous self-supervision tasks, our objective is to design a simple and uniform self-supervision task that can realize both of these two types of tasks.
Moreover, we hope that our self-supervision task can incentivize the representation to capture the long-term dynamics instead of only predicting the one-step dynamics.

Accordingly, we propose a self-supervision task that predicts the masked states or actions in the trajectory.
We present our pre-training model in Figure~\ref{fig:model} and introduce the pre-training model as follows:

\textbf{Mask.}
Given a trajectory $(s_{1}^i, a_1^i, s_2^i, a_2^i, \cdots, s_T^i)$ in the offline dataset, we first randomly select an anchor item (i.e., state or action) in the sequence and mask several consecutive items of the same kind starting from this anchor item. 
Each item can be selected as the anchor with probability $p$ and the number of consecutive masked items is sampled from $\text{Unif}(n)$ (a discrete uniform distribution over $1\sim n$), where $p\in\mathbb{R}$ and $n\in\mathbb{Z}$ are hyperparameters.
Taking the case in Figure~\ref{fig:model} as an example, we select the state $s_t$ and mask $2$ consecutive states starting from this state (i.e., $s_t$ and $s_{t+1}$).
Similar to BERT \citep{devlin2018bert}, we replace the masked elements with 1)~a fixed token 80\% of the time; 2)~an element of the same kind randomly sampled in the dataset 10\% of the time; and 3)~the unchanged element 10\% of the time.
Each trajectory typically contains thousands of steps, which can lead to a large amount of computational cost for the self-attention module.
To reduce computational cost, we randomly cut the trajectory into several segments, each of which contains up to $K$ states.

\textbf{Architecture.}
We train the encoder $f(\cdot)$ and the transformer blocks $\text{Trans}
(\cdot)$ to predict the masked elements.
We maintain two different encoders for states and actions respectively.
The states in our later experiments are image-based, and we use a ResNet \citep{he2016deep} architecture as a state encoder.
Notice that, compared with the encoder architecture used for image-based states in popular RL models (such as \citep{liu2020unsupervised,stooke2021decoupling}) with only a few convolution layers, we use a larger encoder architecture.
This is motivated by the recent observation that a larger network architecture can incorporate richer information and is more suitable for the pre-training setting than the online setting in RL \cite{schwarzer2021pretraining}.
Next, we use a two-layer MLP as the action encoder to generate the action representation based on the raw action.
For discrete actions, the raw actions are represented as one-hot vectors.
We use an improved version of transformer blocks called GTrXL \citep{parisotto2020stabilizing} that reorders the identity mapping and additionally adopts a gating layer based on the original transformer structure \citep{vaswani2017attention}.
GTrXL improves the stability and learning speed of the original transformer structure and is shown to be suitable for RL.
We use $L$ GTrXL blocks in our model.

\textbf{Self-supervision task.}
Different from the masked prediction task in BERT \citep{devlin2018bert} where the targets are discrete, our model should predict image-based states.
Therefore, we use the transformer blocks to predict the embeddings of the image-based states or actions instead of the original ones.
We design a contrastive loss based on SimCLR \citep{chen2020simple} for the self-supervision task as follows:
Given a segment of $K$ steps (corresponding to a sequence of length $2K-1$), we denote the index set of masked items as $\mathcal{T}$.
Consider a masked item $\tilde{x}_t$ with index $t\in\mathcal{T}$ that is generated from the original item $x_t$. 
The item $x$ can be the state $s$ or the action $a$.
We denote the embedding of the masked item (i.e., the output of the encoder) as $\tilde{y}_t:=f(\tilde{x}_t)$ for some $t\in\mathcal{T}$ and the embedding of the original item as $y_t:=f(x_t)$ for some $t\in[2T-1]$.
We feed these embeddings to the GTrXL blocks, i.e., $z_1, \cdots, \tilde{z}_t, \cdots, z_K := T(y_1, \cdots, \tilde{y}_t, \cdots, y_K)$, and obtain the predicted embedding for the masked item $\tilde{z}_t$ for some $t\in\mathcal{T}$.
Our aim is to maximize the similarity between the predicted embedding $\tilde{z}_t$ and the target embedding $y_t$ that is the representation of the original item $x_t$.
To better calculate the similarity of two high-dimensional vectors, we use a projection function $g(\cdot)$ to map the embeddings to lower-dimensional vectors.
The loss of our self-supervision task is defined as:
\begin{equation}
\scalemath{0.85}{
\begin{aligned}
\mathcal{L}:= - \frac{1}{|\mathcal{T}|} \sum_{t\in\mathcal{T}} 
\Bigg[ &
\log \dfrac{\exp\big( \text{sim} (g(\tilde{z}_t), g(y_t)) \big)}{ \sum\limits_{\tau\in[2K-1] } \mathbb{I}[\tau \neq t] \exp\big( \text{sim} (g(\tilde{z}_{t}), g(y_{\tau})) \big) } \\
& +
\log \dfrac{\exp\big( \text{sim} (g(y_t),g(\tilde{z}_t)) \big)}{ \sum\limits_{\tau\in[2K-1] } \mathbb{I}[\tau \neq t] \exp\big( \text{sim} (g(y_{t}), g(z_{\tau})) \big) }
\Bigg],
\end{aligned}
}
\end{equation}
where the cosine similarity is defined as $\text{sim}(z_1, z_2):=z_1^Tz_2/(\Vert z_1 \Vert \Vert z_2 \Vert)$.

\textbf{Discussion.}
This self-supervision task consists of two types of sub-tasks depending on the type of the masked items: the state prediction task and the action prediction task.
When the model tries to predict the first several states in a sequence of masked states, it essentially predicts the future states based on the past states and actions.
This is similar to the forward prediction task in previous papers.
When the model tries to predict the masked action, this corresponds to the previous backward prediction task.
Moreover, our self-supervision task emphasizes more on predicting based on the sequence instead of single states or actions, which is more suitable for the partially observable setting where the transition cannot be modeled accurately using the one-step transition.

\subsection{Downstream RL Tasks}

In specific RL tasks, we only use the state encoder learned in the pre-training phase.
Notice that the action encoder also incorporates useful prior knowledge, and it should be beneficial to utilize it in specific RL tasks, but we leave it as a future research direction.
Broadly speaking, there are two approaches to utilizing the pre-trained encoder: fine-tuning and using as fixed features.
In our experiments, we find that using fixed features works better for RL. 
This has also been validated in previous papers such as \citep{seo2022reinforcement} in which the authors explain that the fine-tuning scheme can quickly erase the useful knowledge in the encoder.
Therefore, we fix the state encoder and append several additional MLP layers as the value network, the policy network, the network for dynamic prediction, etc., according to different downstream tasks.



\section{Experiments}
\label{sec:experiments}

We design our experiments to answer the following questions:
\begin{itemize}
    \item (Q1, Atari) Can RePreM improve the performance of value-based RL algorithms on the Atari-100k benchmark compared with previous algorithms?
    \item (Q2, Atari) How does the performance of RePreM change with the size of the dataset, the quality of the dataset, and the size of the state encoder?
    \item (Q3, Atari) Can the representations generated by RePreM be used for dynamic prediction and capture the long-term dynamics better than the previous algorithms? 
    \item (Q4, Atari) Can the representations generated by RePreM be transferred to accelerate the learning for unseen tasks?
    \item (Q5, DMControl) Can RePrem also improve the performance of actor-critic algorithms on the DMControl-100k benchmark compared with previous algorithms?
    \item (Q6, Ablation) What is the contribution of each proposed design in RePreM?
\end{itemize}

\begin{table*}[t]
    \centering
    \small
    \begin{tabular}{c c c r r r r}
    No. & Dataset & Methods & Median & Mean & \textgreater H & \textgreater 0  \\
    \hline
    $a1$ & \multirow{5}{3cm}{\centering --} & 
    Rainbow & -0.090 & 0.000 & 0 & 11 \\
    $a2$ && SimPLe & 0.144 & 0.443 & 2 & 26 \\ 
    $a2$ && DER & 0.161 & 0.285 & 2 & 26 \\
    $a3$ && DrQ & 0.268 & 0.357 & 2 & 24 \\
    $a4$ && SPR & 0.415 & 0.704 & 7 & 26 \\
    \hline
    $b1$ & Weak (3M) & ATC
    & 0.219 & 0.587 & 4 & 26 \\
    $b2$ & Mixed (3M) & ATC
    & 0.204 & 0.780 & 5 & 26 \\
    $b3$ & Mixed (3M) & BC w/o fine-tune
    & 0.139 & 0.227 & 0 & 23 \\
    $b4$ & Mixed (3M) & BC
    & 0.548 & 0.858 & 8 & 26 \\
    $b5$ & Weak (5M) & SGI
    & 0.589 & 1.144 & 8 & 26 \\
    $b6$ & Mixed (3M) & SGI (Small)
    & 0.423 & 0.914 & 8 & 26 \\
    $b7$ & Mixed (3M) & SGI 
    & 0.679 & 1.149 & 9 & 26 \\
    $b8$ & Mixed (6M) & SGI (Large)
    & 0.753 & 1.598 & 9 & 26 \\
    \hline
    $c1$ & Random (10M) & RePreM & 0.190 & 0.289 & 4 & 26 \\
    $c2$ & Weak (10M)
    & RePreM & 0.774 & 1.335 & 9 & 26 \\
    $c3$ & Mixed (3M)
    & RePreM & 0.717 & 1.298 & 9 & 26 \\
    $c4$ & Mixed (5M)
    & RePreM & 0.788 & 1.436 & 9 & 26 \\
    $c5$ & Mixed (10M) & RePreM (Small) & 0.835 & 1.384 & 10 & 26 \\
    $c6$ & Mixed (10M) & RePreM & 0.953 & 1.970 & 13 & 26 \\
    $c7$ & Mixed (10M) & RePreM (Large) & \textbf{1.085} & \textbf{2.038} & 15 & 26 \\
    \end{tabular}
    \caption{The performance of different algorithms on Atari-100K. The performance in group $a$ and $b$ are quoted from previous papers or calculated based on the scores provided in previous papers. The performance in group $c$ are the mean across 10 random seeds and 100 evaluation trajectories.}
    \label{tab:main_table}
\end{table*}

\textbf{Dataset.}
We conduct the following experiments on Atari \citep{bellemare2013arcade} and DMControl \citep{tassa2018deepmind}.
For Atari, as in many previous papers (e.g., \citep{kaiser2019model}), we select a set of 26 games.
For pre-training on Atari games, we collect the following three types of datasets with different qualities for each game:
The \emph{Random} dataset is collected by executing uniformly randomly sampled actions at a number of consecutive steps sampled from a Geometric distribution with $p=\frac{1}{3}$.
The \emph{Weak} dataset is collected from the first 1M transitions generated by DQN.
The \emph{Mixed} dataset is obtained by concatenating multiple checkpoints evenly throughout the training of DQN.
The quality of the dataset increases from \emph{Random} to \emph{Mixed}.
We also evaluate the algorithms on datasets of different sizes.
Larger datasets can be obtained by running the above procedure multiple times with different random seeds.
For DMControl, we collect the offline dataset in a similar way to the procedure to collect the \emph{Mixed} dataset using SAC \citep{haarnoja2018soft}. 



\textbf{Baselines algorithms.}
In our experiments, we compare our algorithm with a wide range of previous algorithms including both sample-efficient RL algorithms and pre-training algorithms for RL.

For Atari games, we incorporate the representation pre-trained by RePreM with Rainbow \citep{hessel2018rainbow} in downstream tasks (except for dynamic prediction).
The baseline algorithms include not only sample-efficient online RL algorithms (such as Rainbow, SimPLe \citep{kaiser2019model}, data-effecient Rainbow/DER \citep{van2019use}, DrQ \citep{yarats2020image}, 
and SPR \citep{schwarzer2020data})
but also pre-training RL methods (such as ATC \citep{stooke2021decoupling}, SGI \citep{schwarzer2021pretraining}, and behavior cloning/BC which is shown to be a strong baseline in \citep{schwarzer2021pretraining}).

For DMControl, we incorporate the representation pre-trained by RePreM with SAC \citep{haarnoja2018soft} into downstream tasks.
We compare our algorithm with SAC without pre-training and APV \citep{seo2022reinforcement} which is a state-of-the-art pre-training method.
This method beats many strong sample-efficient RL algorithms, including DrQ \citep{yarats2020image} and Dreamer-v2 \citep{hafner2020mastering}.

\subsection{Q1: Performance of RePreM on Atari-100k}

\textbf{Evaluation metrics.}
Atari-100k is a popular benchmark for sample-efficient RL where the agent is allowed to interact with the environment for only 100k steps, equivalent to approximately two hours of human experience. 
We report the evaluation metrics following \citep{schwarzer2021pretraining}.
The metrics include the mean and median human normalized score (HNS) averaging across 10 random seeds and over 100 evaluation trajectories at the end of the training process, the number of games in which the agent achieves super-human performance (\textgreater H) and greater-than-random performance (\textgreater 0).
HNS is calculated as $\frac{\text{AgentScore}- \text{RandomScore}}{\text{HumanScore} - \text{RandomScore}}$ and we use the agent/human scores listed in \citep{badia2020agent57}.

\textbf{Experiment results.}
To answer the first question, we run RePreM on the \emph{Mixed} dataset with 3M samples, which is comparable to the datasets in previous pre-training methods. 
We present the result in Table \ref{tab:main_table}. 
We can observe that RePreM ($c3$ in Table \ref{tab:main_table}) outperforms not only the previous learning-from-scratch sample efficient algorithms ($a1$-$a5$ in Table \ref{tab:main_table}) but also recently proposed pre-training methods ($b2$-$b4$, $b6$-$b7$ in Table \ref{tab:main_table}).
We also note that the superior performance of RePreM is achieved using Rainbow (which is a popular model-free algorithm) as the base RL algorithm, but the performance of this base algorithm alone is the worst among all the listed algorithms (cf. $a1$ in Table \ref{tab:main_table}).
This indicates that the representation pre-trained by RePreM can turn Rainbow into a strong algorithm.
The results of this experiment indicate that the representation pre-trained by RePreM enables sample-efficient RL effectively.

\subsection{Q2: Scalability of RePreM}

\textbf{Variants of RePreM.}
To evaluate the scalability of RePreM in terms of the dataset size, the dataset quality, and the scale of the encoder, we obtain several variants of the representations pre-trained by RePreM.
First, we use the datasets of 3M, 5M, and 10M samples, respectively, to evaluate the scalability w.r.t. the dataset size.
Second, we use the \emph{Random}, \emph{Weak}, and \emph{Mixed} datasets to evaluate the scalability w.r.t. the quality of the dataset.
Third, we vary the size of the ResNet used in the encoder to evaluate the scalability w.r.t. the scale of the encoder, resulting in \emph{Small}, \emph{Medium} (default), and \emph{Large} variants.

\textbf{Experiment results.}
To evaluate how performance changes with the quality of the dataset, we can compare the performance of RePreM on \emph{Random} (10M), \emph{Weak} (10M) and \emph{Mixed} (10M) (see $c1$, $c2$, and $c6$ in Table \ref{tab:main_table} respectively).
On the basis of these results, we conclude that the performance of RePreM scales with the quality of the datasets.
Moreover, although the performance worsens when the data quality degenerates, RePreM from \emph{Weak} (see $c2$) still outperforms the previous pre-training methods that learn from \emph{Weak} datasets (see $b1$ and $b5$).

To evaluate how performance varies with the size of the dataset, we can compare the performance of RePreM learning from the \emph{Mixed} dataset with 3M, 5M, and 10M samples (see $c3$, $c4$, and $c6$ in Table \ref{tab:main_table} respectively).
We can observe that the performance also scales with the dataset size, which indicates that our method can benefit from a larger dataset.

To evaluate how performance varies with the size of the state encoder, we can observe the performance of RePreM (\emph{Small/Medium/Large}) learned from \emph{Mixed} (10M) (see $c5$-$c7$ in Table \ref{tab:main_table}).
In contrast to the online RL setting, where a large network structure does not lead to performance improvement, RePreM can benefit from a larger network structure.

Finally, combined with a large moderate-quality dataset (\emph{Mixed} 10M) and a large state encoder, RePreM can achieve a performance that significantly outperforms the other algorithms and variants.
The scaling trends of RePreM indicate that the performance of RePreM should further increase with the scale of parameters and datasets. 

\subsection{Q3: RePreM for Dynamic Prediction}
\label{sec:dynamic_prediction}

\begin{figure}
    \centering
    \includegraphics[width=0.4\textwidth]{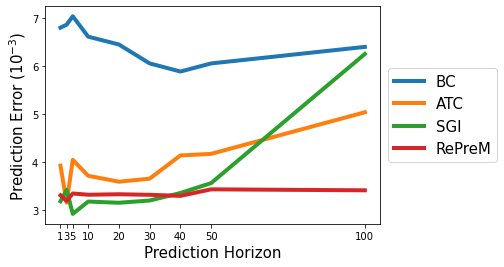}
    \caption{The prediction error versus different prediction horizons on Alien. We fix the pre-trained representations from different algorithms and append a trainable prediction network (i.e., a two-layer MLP) to predict the future RAM state. 
    We train the prediction network with 10 random seeds and show the mean.}
    \label{fig:long_term}
\end{figure}

We evaluate the representations learned by different pre-training algorithms with the dynamic prediction task that predicts the future states based on the learned representation on Atari games.
Specifically, we fix the state encoder pre-trained with different methods and append two MLP layers to the encoder.
We train this network to predict the future RAM state $\bar{s}_{t+k}$ (divided by 255) based on the current state $s_t$.
We collect the training and testing samples using the same DQN policy and evaluate the prediction error using the mean-squared error between the predicted RAM state and the ground truth.
We present the results obtained on Alien in Figure \ref{fig:long_term} and the results are similar across different games.
We can observe that the representations generated by RePreM and SGI result in a more accurate prediction for the states in the near future than those of BC and ATC. 
This may be due to the fact that RePreM and SGI use forward prediction as one of the self-supervision tasks.
Moreover, we find that, compared with SGI, the model based on RePreM can predict the states in the far future with higher accuracy.
This indicates that RePreM generates representations that can better capture long-term information.
This may explain the reason why RePreM outperforms the other representation pre-training algorithms in the Atari-100k benchmark.

\subsection{Q4: Transfer to Unseen Tasks}

\begin{figure}[t]
    \centering
    \includegraphics[width=\columnwidth]{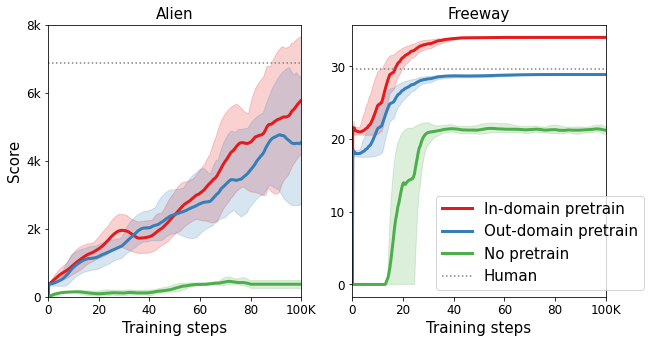}
    \caption{Comparison on the performance between the RePreM representation pre-trained on the same task and that pre-trained on the remaining 24 tasks. The shaded area indicates the standard deviation over 10 random seeds.}
    \label{fig:transfer}
\end{figure}

To evaluate how the representations pre-trained by RePreM can transfer the knowledge to unseen tasks, we select Alien and Freeway as the target tasks and pre-train the representation using RePreM on a dataset collected on the 24 remaining Atari games with 12M samples (with 0.5M samples from the \emph{Mixed} dataset for each game).
The pre-trained representation is used in the same way as Atari-100k.
We compare it with learning curves from the representation pre-trained on the homogeneous task (with 10M \emph{Mixed} samples) and learning-from-scratch using the same network structure. 
We present the results in Figure \ref{fig:transfer}. 
The result indicates that the representation pre-trained by RePreM can extract the knowledge transferable to other tasks.

\subsection{Q5: Performance of RePreM on DMC-100k}

We first collect the offline dataset by SAC on the Walker and Quadruped environments from DMControl. Then, we pre-train the representation using RePreM and run SAC based on the pre-trained representation.
The pre-trained representation is used as the encoder of both the critic and the actor in SAC.
We compare RePreM with SAC from scratch and APF.
For fair comparison, instead of using the original APF model pre-trained on different domains, we pre-train their model using the same dataset as ours and this improves the performance over the version provided in their paper.

We show the performance in Table \ref{tab:dmc}.
First, we can see that SAC based on the RePreM representation performs much better than learning from scratch, which indicates the effectiveness of the pre-trained representation.
Second, we note that, different from RePreM that only uses the pre-trained representation, APF uses not only the pre-trained representation but also the latent transition model and the image decoder.
Nevertheless, with the representation pre-trained by RePreM, SAC achieves comparable performance with this state-of-the-art pre-training method.
In addition, note that we use the same RePreM representation for these tasks, and RePreM+SAC achieves good performance. 
This indicates that the RePreM representation can work for downstream tasks with different reward functions.


\begin{table}[t]
    \centering
    \begin{tabular}{lrrrrr}
    Task & SAC & APV & RePreM \\
    \hline
    Walker-Walk & 42$\pm$12 & 444$\pm$36 & \textbf{633}$\pm$52 \\
    Walker-Run & 67$\pm$27 & 496$\pm$74 & \textbf{530}$\pm$43 \\
    Walker-Stand & 142$\pm$21 & 742$\pm$81 & \textbf{899}$\pm$38 \\
    Quadruped-Walk & 127$\pm$184 & 420$\pm$87 & \textbf{799}$\pm$132 \\
    Quadruped-Run & 93$\pm$36 & \textbf{698}$\pm$96  & 579$\pm$85 \\
    \end{tabular}
    \caption{The performance of different algorithms on Walker from DMControl-100K.}
    \label{tab:dmc}
\end{table}

\subsection{Q6: Ablation Studies}

To evaluate the contribution of each proposed design, we implement the following ablated versions on top of the RePreM version in $c3$ of Table~\ref{tab:main_table}.
The results are presented in Table~\ref{tab:ablation}.
\begin{itemize}
    \item \emph{RePreM+FT} that fine-tunes the network in the downstream tasks instead of fixing the pre-trained state encoder.
    We can see that fine-tuning in downstream tasks degenerates the performance.
    This result is consistent with \citep{seo2022reinforcement} who finds that fine-tuning may quickly lose the pre-trained representation.
    \item \emph{RePreM+short} that models shorter sequences with $T=10$ (instead of $T=50$ in RePreM).
    The experiment result shows that using a shorter sequence length significantly degenerates the performance.
    This may indicate that capturing the long-term dynamics with a large $T$ is important to the success of RePreM.
    \item \emph{RePreM-BERT} that replaces all masked elements with a fixed token instead of following the masking scheme in BERT.
    The result indicates that following the masking scheme in BERT also contributes to the success of RePreM.
    \item \emph{RePreM-GTrXL} that uses the original transformer block instead of GTrXL. We can see that using GTrXL can improve performance, possibly because of its stability.
    \item \emph{RePreM+ReLIC} that uses the ReLIC \citep{mitrovic2020representation} loss as the contrastive self-supervision loss instead of the SimCLR loss for the masked prediction (as in CoBERL \citep{banino2021coberl}). We can see that the performance of the simple SimCLR loss is similar to that of the ReLIC loss, but ReLIC is more complex. For simplicity, we use SimCLR in our proposed method.
    \item \emph{RePreM+decoder} that predicts the raw states/actions with state/action decoders instead of predicting their embeddings using the SimCLR loss (as in \citep{jaderberg2016reinforcement,seo2022reinforcement}).
    We can see that using the encoder to predict the raw states/actions underperforms the proposed RePreM that predicts the embeddings significantly. 
    The reason may be that the raw states are high-dimensional, and reconstructing the raw states may mislead the encoder and the transformer to learn spurious details.
\end{itemize}

\begin{table}[t]
    \centering
    \begin{tabular}{crrrrr}
    Ablated Versions & Median & Mean & \textgreater H & \textgreater 0 \\
    \hline
    RePreM+FT & 0.512 & 0.839 & 7 & 26 \\
    RePreM+short & 0.322 & 0.806 & 5 & 26 \\
    RePreM-BERT & 0.588 & 1.166 & 8 & 26 \\
    RePreM-GTrXL & 0.709 & 1.220 & 9 & 26 \\
    RePreM+ReLIC & 0.715 & 1.201 & 9 & 26 \\
    RePreM+decoder & 0.107 & 0.210 & 0 & 24 \\
    RePreM ($c3$ in Table \ref{tab:main_table}) & {0.717} & {1.298} & 9 & 26 \\
    \end{tabular}
    \caption{The performance of different ablated variants on Atari-100K pre-trained using the Mixed dataset with 3M samples. These ablated versions use the medium-sized ResNet as the state encoder.}
    \label{tab:ablation}
\end{table}

\section{Conclusion}
\label{sec:conclusion}

In this paper, we propose a representation pre-training method called RePreM (Representation Pre-training with Masked Model) for RL.
RePreM is algorithmically simple but effective, through learning the representation with only a transformer-based sequence model and masked prediction.
Although RePreM adopts sequence modeling that is different from the previous representation pre-training baselines for RL, it performs on a par with or better than previous RL pre-training methods.
Empirically, we find that the success of RePreM can be attributed to its strong ability to capture long-term dynamics in the trajectories.
This is a benefit of modeling the whole sequence instead of one-step dynamics as in previous methods.
Through extensive ablation studies, we show that the representation pre-trained using RePreM can be used for various downstream tasks.
We also find that RePreM not only scales well with the dataset size, the dataset quality, and the size of the encoder, but can also generalize across different tasks.
This indicates its potential to extract knowledge from big data and use the knowledge for efficient and robust learning.
Currently, RePreM only uses the state encoder learned in the pre-training phase but discards the transformer blocks and action encoder that also contain useful knowledge. 
Thus, how to utilize the knowledge in the transformer blocks and action encoder will be an interesting future research direction.

\section*{Acknowledgement}

The work of Yuanying Cai and Longbo Huang is supported in part by the Technology and Innovation Major Project of the Ministry of Science and Technology of China under Grant 2020AAA0108400 and 2020AAA0108403, the Tsinghua University Initiative Scientific Research Program, and Tsinghua Precision Medicine Foundation 10001020109.
Dr. Xuyun Zhang is the recipient of an ARC DECRA (project No. DE210101458) funded by the Australian Government.

\balance
{
\footnotesize
\bibliography{aaai23}
}


\end{document}